\documentclass[letterpaper, 10 pt, conference]{ieeeconf}
\IEEEoverridecommandlockouts
\usepackage{acro}
\let\labelindent\relax
\usepackage{enumitem}

\usepackage{graphics}
\usepackage{graphicx} 
\usepackage{multirow}
\usepackage{longtable}
\usepackage{epsfig}
\usepackage{epstopdf}

\usepackage{amsmath, mathrsfs, dsfont}
\usepackage{amssymb}
\usepackage{amsfonts}
\usepackage{bm}

\usepackage{cite}
\usepackage{verbatim}

\usepackage{amscd}
\usepackage{color}
\definecolor{lgray}{gray}{0.6}
\usepackage{rotating}

\usepackage[font=footnotesize]{subcaption}

\usepackage{mathptmx}
\usepackage[11pt]{moresize}
\usepackage{flushend}
\usepackage[stable]{footmisc}

\usepackage{algorithmicx}
\usepackage{algorithm}
\usepackage{algpascal}
\usepackage{float}
\usepackage{algc}
\usepackage{algcompatible}
\usepackage{algpseudocode}

\usepackage[vcentermath]{youngtab} 
\usepackage{url}

\newcommand{\BoldB}				{ \mathbf{B} }
\newcommand{\Boldb}				{ \mathbf{b} }

\newcommand{\Boldd}				{ \mathbf{d} }

\newcommand{\Bolde}				{ \mathbf{e} }

\newcommand{\Boldg}				{ \mathbf{g} }
\newcommand{\BoldG}				{ \mathbf{G} }

\newcommand{\BoldH}				{ \mathbf{H} }
\newcommand{\Boldh}				{ \mathbf{h} }

\newcommand{\BoldJ}				{ \mathbf{J} }

\newcommand{\BoldP}				{ \mathbf{P} }
\newcommand{\Boldp}				{ \mathbf{p} }
\newcommand{\BoldQ}				{ \mathbf{Q} }
\newcommand{\Boldq}				{ \mathbf{q} }
\newcommand{\BoldR}				{ \mathbf{R} }

\newcommand{\Boldu}				{ \mathbf{u} }

\newcommand{\Boldv}				{ \mathbf{v} }
\newcommand{\Boldw}				{ \mathbf{w} }

\newcommand{\Boldx}				{ \mathbf{x} }

\newcommand{\Boldz}				{ \mathbf{z} }

\newcommand{\Rbb}			    { \mathbb{R} }

\newcommand{\0}					{ \boldsymbol{0} }
\newcommand{\1}					{ \boldsymbol{1} }

\newcommand{\Boldomega}			{ \boldsymbol{\omega} }

\newcommand{\Boldepsilon}			{ \boldsymbol{\epsilon} }

\newcommand{\Boldgamma}			{ \boldsymbol{\gamma} }

\newcommand{\BoldPhi}			{ \boldsymbol{\Phi} }

\newcommand{\Boldmu}			{ \boldsymbol{\mu} }

\newcommand{\Boldchi}			{ \boldsymbol{\chi} }
\newcommand{\BoldPsi}           {\boldsymbol{\Psi}}

\newcommand\norm[1]{\lVert#1\rVert}

\newcommand\T{\rule{0pt}{2.6ex}}       
\newcommand\B{\rule[-1.2ex]{0pt}{0pt}} 

\newcounter{inlineenum}
\renewcommand{\theinlineenum}{\alph{inlineenum}}

\newcommand{\brmrk}[1]{\begin{remark} \label{#1} }
	\newcommand{\ermrk}{ \hfill $\bigtriangleup$    \end{remark} \vspace{1mm} }

\newtheorem{exercise}{Exercise}[section]
\newcommand{\boex}[1]{\begin{example} \label{#1} --- \rm}
	\newcommand{\eoex}{ \hfill $\bigtriangleup$    \end{example} \vspace{1mm} }
\newtheorem{example}{Example}[section]
\newcommand{\bohw}[1]{\begin{exercise} \label{#1} -- \rm}
	\newcommand{\eohw}{ \hfill    \end{exercise} \vspace{1mm} }
\newtheorem{assumption}{Assumption}[section]
\newcommand{\boass}[1]{\begin{assumption} \label{#1} -- \rm}
	\newcommand{\eoass}{ \hfill    \end{assumption} \vspace{1mm} }

\renewcommand{\baselinestretch}{1.0}

\newcommand{\black}{\color{black}}

\definecolor{brinkpink}{rgb}{1.00, 0.33, 0.64}



\usepackage{tikz}
\usepackage{wrapfig}
\usepackage{stfloats}
\DeclareMathAlphabet\mathbfcal{OMS}{cmsy}{b}{n}
\DeclareAcronym{APC}{
	short=APC,
	long=Antenna Phase Center,
}

\DeclareAcronym{COM}{
short=COM,
long=Center of Mass,
}

\DeclareAcronym{CV}{
short=CV,
long=Connected Vehicle,
}

\DeclareAcronym{CAV}{
	short=CAV,
	long=Connected Automated Vehicles,
}

\DeclareAcronym{DCB}{
short=DCB,
long=Differential Code Bias,
}

\DeclareAcronym{DSRC}{
	short=DSRC,
	long=Dedicated Short-Range Communications,
}

\DeclareAcronym{CDMA}{
	short=CDMA,
	long=Code Division Multiple Access,
}

\DeclareAcronym{CDF}{
	short=CDF,
	long=Cumulative Distribution Function,
}

\DeclareAcronym{CE-CERT}{
	short=CE-CERT,
	long=College of Engineering Center for Environmental Research and Technology,
}

\DeclareAcronym{CME}{
	short=CME,
	long=common-mode errors,
}

\DeclareAcronym{FDMA}{
	short=FDMA,
	long=Frequency Division Multiple Access,
}

\DeclareAcronym{DD}{
	short=DD,
	long=Double-Differenced,
}

\DeclareAcronym{DGPS}{
	short=DGPS,
	long=Differential Global Positioning System,
}

\DeclareAcronym{DLR}{
	short=DLR,
	long=German Aerospace Center,
}
\DeclareAcronym{DGNSS}{
short=DGNSS,
long=Differential GNSS,
}
\DeclareAcronym{ECEF}{
short=ECEF,
long=Earth-Centered Earth-Fixed,
}
\DeclareAcronym{ECI}{
	short=ECI,
	long=Earth-Centered Inertial,
}
\DeclareAcronym{KF}{
	short=KF,
	long=Kalman Filter,
}
\DeclareAcronym{EKF}{
	short=EKF,
	long=Extended KF,
}
\DeclareAcronym{GPS}{
	short=GPS,
	long=Global Positioning System,
}
\DeclareAcronym{GNSS}{
short=GNSS,
long=Global Navigation Satellite Systems,
}
\DeclareAcronym{GPSSPS}{
short=GPS SPS,
long=GPS standard positioning service,
}

\DeclareAcronym{HMM}
{short = HMM, long = Hidden Markov Model}

\DeclareAcronym{HD}
{short = HD, long = Horizontal Distance}

\DeclareAcronym{HiD}{
	short=Hi-Def,
	long=High-definition,
}

\DeclareAcronym{IMU}{
short=IMU,
long=Inertial Measurement Unit,
}

\DeclareAcronym{INS}{
	short=INS,
	long=Inertial Navigation System,
}

\DeclareAcronym{ILP}{
short=ILP,
long=Integer Linear Programming,
}

\DeclareAcronym{I2V}{
short=I2V,
long=Infrastructure-to-Vehicle,
}
\DeclareAcronym{IGS}{
	short=IGS,
	long=International GNSS Service,
}
\DeclareAcronym{ISB}{
	short=ISB,
	long=inter-system bias,
}

\DeclareAcronym{LLMM}
{short = LLMM, long = Lane-level Map-matching}

\DeclareAcronym{LD}
{short = LD, long = Lane Determination}

\DeclareAcronym{LOS}
{short = LOS, long = line-of-sight}

\DeclareAcronym{NLOS}
{short = NLOS, long = non-line-of-sight}

\DeclareAcronym{LMI}
{short = LMI, long = Linear Matrix Inequality}

\DeclareAcronym{RLMM}
{short = RLMM, long = Road-level Map-matching}

\DeclareAcronym{MSE}
{short = MSE, long = Mean Square Error}

\DeclareAcronym{MGEX}
{short = MGEX, long = Multi-GNSS Experiment}

\DeclareAcronym{MAP}
{short = MAP, long = Maximum-a-Posteriori}
\DeclareAcronym{OS}{
	short=OS,
	long=Open Service,
}

\DeclareAcronym{OSB}{
short=OSB,
long=Observable-specific Code Biases,
}
\DeclareAcronym{OSR}{
short=OSR,
long=Observation Space Representation,
}
\DeclareAcronym{PPP}{
	short=PPP,
	long=Precise Point Positioning,
} 

\DeclareAcronym{RTPPP}{
	short=RT-PPP,
	long=Real-time PPP,
}

\DeclareAcronym{PPP-AR}{
short=PPP-AR,
long=Precise Point Positioning Ambiguity Resolution,
} 

\DeclareAcronym{PP}{
short=PP,
long=Post Processing,
}

\DeclareAcronym{PVA}{
	short=PVA,
	long={position, velocity, acceleration},
} 

\DeclareAcronym{RTCM}{
short=RTCM,
long=Radio Technical Commission for Maritime Services,
}
\DeclareAcronym{RTK}{
short=RTK,
long=Real-time Kinematic,
}
\DeclareAcronym{SBAS}{
short=SBAS,
long=Satellite Based Augmentation Systems,
}

\DeclareAcronym{SNR}{
short=SNR,
long=Signal-to-Noise Ratio,
}
\DeclareAcronym{SSR}{
short=SSR,
long=State Space Representation,
}
\DeclareAcronym{SPS}{
	short=SPS,
	long=Standard Positioning Service,
}

\DeclareAcronym{SAE}{
	short=SAE,
	long=Society of Automotive Engineers,
}

\DeclareAcronym{STD}{
	short=STD,
	long=Standard Deviation,
}

\DeclareAcronym{STEC}{
	short=STEC,
	long=Slant Total Electron Content,
}
\DeclareAcronym{VTEC}{
	short=VTEC,
	long=Vertical Total Electron Content,
}

\DeclareAcronym{SH}{
	short=SH,
	long=Spherical Harmonic,
}
\DeclareAcronym{TGD}{
short=TGD,
long=Timing Goup Delay,
}
\DeclareAcronym{TD}{
	short=TD,
	long=Threshold Decisions,
}
\DeclareAcronym{TOP}{
	short=TOP,
	long=time-of-signal-propagation,
}
\DeclareAcronym{TOT}{
	short=TOT,
	long=time-of-signal-transmission,
}
\DeclareAcronym{TOR}{
	short=TOR,
	long=time-of-signal-reception,
}
\DeclareAcronym{ZTD}{
short=ZTD,
long=Zenith Troposphere Delay,
}
\DeclareAcronym{TEC}{
short=TEC,
long=Total Electron Content,
}
\DeclareAcronym{IPP}{
	short=IPP,
	long=Ionosphere Pierce Point,
}
\DeclareAcronym{NOAA}{
	short=NOAA,
	long=National Oceanic and Atmospheric Administration,
}
\DeclareAcronym{UCR}{
	short=UCR,
	long=University of California-Riverside,
}
\DeclareAcronym{USTEC}{
short=US-TEC,
long=US Total Electron Content,
}
\DeclareAcronym{VNDGNSS}{
	short=VN-DGNSS,
	long=Virtual Network DGNSS,
}
\DeclareAcronym{IOD}{
	short=IOD,
	long=Issue Of Data,
}
\DeclareAcronym{CAS}{
	short=CAS,
	long=Chinese Academy of Sciences,
}
\DeclareAcronym{CNES}{
	short=CNES,
	long=Centre National d'Études Spatiales,
}
\DeclareAcronym{RMS}{
	short=RMS,
	long=Root Mean Square,
}
\DeclareAcronym{SF}{
	short=SF,
	long=Single Frequency,
}

\DeclareAcronym{DF}{
	short=DF,
	long=Dual Frequency,
}
\DeclareAcronym{ICD}{
	short=ICD,
	long=Interface Control Document,
}
\DeclareAcronym{NED}{
	short=NED,
	long={North, East and Down},
}
\DeclareAcronym{WAAS}{
	short=WAAS,
	long=Wide Area Augmentation System,
}
\DeclareAcronym{OPUS}{
	short=OPUS,
	long=Online Positioning User Service,
}
\DeclareAcronym{PDF}{
	short=PDF,
	long=Probability Density Function,
}
\DeclareAcronym{RAPS}{
	short=RAPS,
	long=Risk-Averse Performance-Specified,
}

\DeclareAcronym{VRS}{
	short=VRS,
	long=Virtual Reference Station,
}

\DeclareAcronym{SPP}{
	short=SPP,
	long=Single-frequency Point Positioning,
}

\DeclareAcronym{SPaT}{
	short=SPaT,
	long=Signal Phase and Timing,
}

\DeclareAcronym{TECU}{
	short=TECU,
	long=Total Electron Content Units,
}

\DeclareAcronym{BNC}{
	short=BNC,
	long=BKG NTRIP Client,
}

\DeclareAcronym{ITS}{
	short=ITS,
	long=Intelligent Transportation Systems
}

\DeclareAcronym{USDOT}{
	short=USDOT,
	long=U.S. Department of Transportation
}

\DeclareAcronym{WHU}{
	short=WHU,
	long=Wuhan University
}

\usepackage[normalem]{ulem}

\usepackage{hyperref}

\begin{document}
	\title{Optimization-Based Outlier Accommodation for Tightly Coupled RTK-Aided Inertial Navigation Systems in Urban Environments}
	
	\author{Wang~Hu,~Yingjie~Hu,~Mike~Stas,
	and~Jay~A.~Farrell
	\thanks{W. Hu (whu027@ucr.edu), M. Stas (mstas001@ucr.edu) and J. A. Farrell (farrell@ece.ucr.edu) are with the Dept. of Electrical and Computer Engineering,
	U. of California, Riverside, CA 92521, USA. Y. Hu (hu000258@umn.edu) is with the Dept. of Aerospace Engineering and Mechanics, U. of Minnesota, Twin Cities.}
	}
	\maketitle

\begin{abstract}
\ac{GNSS} aided \ac{INS} is a fundamental approach for attaining continuously available absolute vehicle position and full state estimates at high bandwidth. 
For transportation applications, stated accuracy specifications must be achieved, unless the navigation system can detect when it is violated.
In urban environments, \ac{GNSS}  measurements are susceptible to  outliers, which motivates the important problem of accommodating outliers while either achieving a performance specification or communicating that it is not feasible. 
\ac{RAPS} is designed to optimally select measurements to address this problem.
Existing  \ac{RAPS} approaches lack a method applicable to carrier phase measurements, which have the benefit of measurement errors at the centimeter level along with the challenge of being biased by integer ambiguities.
This paper proposes a \ac{RAPS} framework that combines \ac{RTK} in a tightly coupled \ac{INS} for urban navigation applications. Experimental results demonstrate the effectiveness of this RAPS-INS-RTK framework, achieving 85.84\% and 92.07\% of horizontal and vertical errors less than 1.5 meters and 3 meters, respectively, using a smartphone-grade \ac{IMU} from a deep-urban dataset. This performance not only surpasses the \ac{SAE} requirements, but also shows a 10\% improvement compared to traditional methods. 
\end{abstract} 	
		
\section{Introduction}\label{sec:intro}
Accurate and reliable real-time localization is essential for automated and connected vehicles in \ac{ITS} \cite{9435134,felipe2023}.
Localization is often achieved through vehicle state estimation by fusing vehicle dynamics across different domains and addressing \ac{INS} uncertainties \cite{xia2021advancing}.
Continuous absolute positioning with high bandwidth often relies on \ac{GNSS} aided \ac{INS} that is achieved through INS error state estimation \cite{9955423,10507228,lipeng2024prioritized}. 
However, in urban environments, \ac{GNSS} measurements are especially susceptible to outliers due, for example, to multipath effects and non-line-of-sight errors. 
The term outlier refers to those measurements that deviates significantly with respect to the model and the rest in the measurement set \cite{barnett1994outliers,ding-24-detection,li-23-deception-detection}. 
Outliers pose substantial challenges to maintaining the accuracy and reliability of navigation solutions \cite{liu2024td3,liu2024enhanced,jin2024learning}.

Traditional methods of outlier rejection in state estimation typically utilize adaptive threshold tests to evaluate the residuals of measurements \cite{patton1991fault,frank1997survey}. 
Specifically, \ac{GNSS} employs Receiver Autonomous Integrity Monitoring (RAIM), where measurement consistency is checked against calculated statistics \cite{anja2017faultDetection}. 
Some advanced approaches like Least Soft-threshold Squares \cite{wang2015robust} and Least Trimmed Squares \cite{rousseeuw2005robust} have been developed for least squares applications. 
Despite their advancements, these methods predominantly focus on detecting and managing outliers, not addressing the trade-offs between risk and performance across the measurements.

An alternative approach to outlier accommodation in state estimation involves optimally selecting a subset of measurements that are consistent with themselves and the prior. 
This method is exemplified by the \ac{RAPS} approach, which minimizes risk while satisfying  performance constraints \cite{Aghapour_TCST_2019,rahman2018outlier,hu2024rapsppp,hu2024optimization}. Prior research has applied the \ac{RAPS} approach to \ac{GNSS}-\ac{INS} systems using GPS-only code measurements in environments with clear sky visibility, where the constraints are consistently feasible and the number of measurements ($m$) is small \cite{Aghapour_TCST_2019}. 
When these approaches utilizes binary measurement selection, its optimization is NP-hard.
Also, its performance constraint using the full information matrix leads to a semi-definite programming problem that becomes runtime unpractical as $m$ increases. 
Recent advancements have introduced a real-time feasible method by employing non-binary variables and a diagonal information matrix in a soft-constrained framework \cite{hu2024rapsppp, hu2024optimization}. 
The non-binary approach, as opposed to a binary inclusion-exclusion model, significantly enhances the computational efficiency of the \ac{RAPS} optimization process \cite{hu2024optimization}. 
The introduction of the soft constraint is aimed at urban environments where performance constraints might not be feasible even with all measurements enabled.  Throughout the rest of this paper, \ac{RAPS} refers to the diagonal form of \ac{RAPS} with a soft constraint.

Previous studies have not evaluated \ac{RAPS} positioning performance using centimeter-accuracy carrier phase measurements (i.e., \ac{RTK}). 
Accurately estimating integer carrier phase ambiguities, a pivotal challenge in \ac{RTK} applications, becomes more complex in urban settings due to frequent signal outages and cycle slips. 
The potential benefits of \ac{RTK} highlight the need for robust methodologies that can effectively evaluate and utilize carrier phase measurements within the urban navigational landscape.

This paper embraces an \ac{RTK} float solution using the instantaneous (single-epoch) mode where float solution aims to offer decimeter-level accuracy.
This mode is uniquely beneficial as it re-initializes carrier phase ambiguities at each epoch, relying solely on single-epoch observations and is thereby immune to cycle slips \cite{parkins2011increasing,odolinski2015combined}. 
This approach circumvents the traditional challenges posed by continuously monitoring for cycle slips.
However, incorporating phase measurements directly into the \ac{RAPS} optimization framework introduces significant complexities. 
In instantaneous \ac{RTK}, there is no prior information about the carrier phase  ambiguities. 
Also, the phase measurements, which have centimeter measurement error levels, can contribute significant information, but appear to be very high risk in the normal \ac{RAPS} approach, even though they are immune from outliers in the instantaneous \ac{RTK} approach. 
Therefore, a new approach is required to accommodate the unique aspects of \ac{GNSS} phase measurements.

This paper introduces a novel \ac{RTK}-\ac{RAPS} framework for outlier accommodation in \ac{RTK}-\ac{GNSS} aided \ac{INS} \footnote{The source code for the experiment is available at \url{https://github.com/Azurehappen/UrbanRTK-INS-OutlierOpt}.}. 
Because phase measurements are less affected by multipath and noise compared to pseudorange, and the ambiguity estimation depends on both the code and phase measurements, our strategy for \ac{RTK}-\ac{RAPS} involves two steps. 
First, this strategy optimally chooses a non-binary measurement selection vector using only the code and Doppler measurements; 
then the code selection is applied to both the code and phase measurement for each satellite.
It then performs the measurement update using the weighted code, phase, and Doppler measurements to achieve the \ac{RAPS} \ac{RTK} float solution.

The experimental analysis section uses the open-source TEX-CUP dataset, acquired within a deep-urban environment \cite{narula2020tex}. 
The results indicate that the \ac{RAPS} \ac{RTK}-aided \ac{INS} approach offers a 10\% improvement in the percentage of epochs satisfying the horizontal and vertical accuracy specifications relative to standard \ac{EKF} and \ac{TD} approaches. 
Notably,  \ac{RAPS} \ac{RTK}-\ac{GNSS} aided \ac{INS}  maintains smoother performance and demonstrates significantly reduced maximum errors.

\black

This paper is organized as follows. 
Sect. \ref{sec:problem} 
reviews the \ac{INS} time propagation, 
the \ac{MAP} approach for measurement updates, and 
the diagonal \ac{RAPS} approach. 
Sect. \ref{sec:estimation} introduces \ac{GNSS} measurement models and proposes a novel method to include \ac{RTK} \ac{GNSS} into an \ac{INS}-\ac{RAPS} integration. 
Sect. \ref{sec:exp} presents an experimental analysis.
Sect. \ref{sec:conclu} conclude this paper.

\section{Problem Background}\label{sec:problem}
The state vector is
\begin{equation} \label{eqn:eqState}
	\Boldx = [\Boldp;\,\Boldv;\,\Boldq_e^b;\,\BoldB_a;\,\BoldB_g] \in \Rbb^n,
\end{equation}
where $\Boldp,\,\Boldv \in \Rbb^3$ represent the \ac{IMU} position and velocity  in the \ac{ECEF} frame; 
$\Boldq_e^b \in \Rbb^4$ denotes the quaternion that parameterizes rotation from the \ac{ECEF} frame to the \ac{IMU} frame; and, 
$\BoldB_a \in \Rbb^3$ and $\BoldB_g \in \Rbb^3$ denote the accelerometer bias and gyroscope bias respectively.
The \ac{IMU} frame is assumed to be coincident with the body frame. 

\subsection{Time Propagation Model and Computation}\label{sect:INS}
The propagation of the state vector through time is described by
\begin{align}\label{eqn:f_true}
	\dot{\Boldx} = f(\Boldx(t),\,\Boldu(t))
\end{align}
where $f()$ represents the known nonlinear kinematic model and $\Boldu \in \Rbb^6$ contains the specific force and angular rate vectors. 
These equations are well-documented and can be found in many references (see e.g., Section 11.2.2 in \cite{farrell2008aided}). 

The \ac{INS} propagates the navigation state vector through time by numerically integrating\footnote{The notation $\circeq$ is used to denote navigation system computations.}
\begin{align}\label{eqn:f_INS}
	\dot{\hat\Boldx} &\circeq f(\hat\Boldx(t),\,\hat\Boldu(t))
\end{align}
where $\hat\Boldx$ is the estimated state vector and $\hat\Boldu(t)\circeq\tilde\Boldu(t)-\hat\BoldB(t)$ is the calibrated IMU measurements. 
The vector $\hat\BoldB=[\hat\BoldB_a ,\,\hat\BoldB_g]\in\Rbb^6$ is the estimated accelerometer and gyroscope bias vector.
The symbol $\tilde\Boldu(t)= \Boldu(t)+\BoldB(t) +\Boldomega(t)$ represents the IMU measurement vector, which is the correct vector $\Boldu$, plus the bias, plus zero mean and white measurement noise $\Boldomega$. 
The error between the true state and the estimated state is
\begin{align}\label{eqn:correction_est}
	\delta \Boldx(t) &= \Boldx(t) -\hat\Boldx(t).
\end{align}
This is a simplified notation, because the attitude error is multiplicative, not additive. 
This topic is  discussed in Appendix D of \cite{farrell2008aided}  and is beyond the scope of this article. 

The IMU samples arrive at a high rate $f_s>100$ Hz.
The GNSS samples arrive at a lower rate rate $f_a\le 1$ Hz.
Therefore, the \ac{INS} integrates eqn. \eqref{eqn:f_INS} over many \ac{IMU} time intervals to time propagate between from measurement epoch $t_{k-1}$ to $t_k$. 
As discussed in Section 7.2.5.2 of \cite{farrell2008aided}, 
the linearized error state propagation model between these two measurement epochs can be accurately represented by
\begin{align}  \label{eqErrStateModel}
		\delta \Boldx_k & = \BoldPhi_k \, \delta \Boldx_{k-1} + \Boldw_k 
\end{align}
where 
$\BoldPhi_k$ is the accumulated state transition matrix and $\Boldw_k \sim \mathcal{N}(\0, \BoldQ_k)$ is the accumulated discrete-time process noise with covariance $\BoldQ_k$.
Based on eqn. \eqref{eqErrStateModel}, state error covariance matrix $\BoldP_{k}\triangleq cov(\delta \Boldx_k)$ time propagation is
\begin{align}  \label{eqtimeUpdate}
		\BoldP^{-}_{k} & \circeq \BoldPhi_k\,\BoldP^{+}_{k-1}\,\BoldPhi^\top_k + \BoldQ_k
\end{align}
The superscripts $-$ and $+$ denote the covariance before and after the aiding  measurement correction. 
Therefore, the prior state estimation error  
$\delta \Boldx_k^- \sim \mathcal{N}(\0, \BoldP_k^-)$.

\subsection{Measurement Update Model and Computation}
The model for the aiding measurement vector $\Boldz_k\in\Rbb^m$ is 
\begin{align}\label{eqn:meas_mdl}
	\Boldz_k = \Boldh(\Boldx) + \Boldgamma_k
\end{align}
where $\Boldh$ is a known function. 
The measurement residual $\delta \Boldz_k\in\Rbb^m$ is computed as
\begin{align}\label{eqn:meas_def}
	\delta \Boldz_k^{-} \circeq \Boldz_k - \Boldh(\hat{\Boldx}^{-}_k).
\end{align}
where $\Boldh(\hat{\Boldx}^{-}_k)$ is the measurement predicted for the state estimate $\hat{\Boldx}^{-}_k$ prior to using the measurement information available at $t=t_k$. 
Substituting eqn. \eqref{eqn:meas_mdl} into  \eqref{eqn:meas_def} and linearizing yields the linear residual model
\begin{align}\label{eqn:lin_meas_mdl}
	\delta \Boldz_k^{-} =  \BoldH_k \, \delta \Boldx_k^{-} + \Boldgamma_k.
\end{align}
where 
$\BoldH_k\circeq \left.\frac{\partial \Boldh(\Boldx)}{\partial \Boldx}\right|_{\Boldx = \hat{\Boldx}^{-}_k}$, and
$\Boldgamma_k \sim \mathcal{N}(\0, \BoldR_k)$ represents measurement noise.
The covariance matrix $\BoldR_k$  is assumed to be invertible and diagonal 
(i.e., $\BoldR_k=diag(\sigma_i^2)$ for $i=1,\cdots,m$).

\subsection{\ac{MAP} Approach}
Given $\delta \Boldx_k^- \sim \mathcal{N}(\0, \BoldP_k^-)$ and  
eqn. \eqref{eqn:lin_meas_mdl} with $\Boldgamma_k \sim \mathcal{N}(\0, \,\BoldR_k)$, minimization of the negative log-likelihood yields 
\begin{align}\label{eqn:map_all}
	\delta \hat{\Boldx}^+_k \circeq \,\underset{\delta \Boldx_k}{\text{argmin}} ~ (\norm{\delta\Boldx_{k}}^2_{\BoldP^-_k} + \norm{\delta \Boldz_k - \BoldH_k \, \delta \Boldx_k}^2_{\BoldR_k}).
\end{align}
The notation  $\norm{\Boldv}^2_{\BoldP}=\Boldv^\top\,\BoldP^{-1}\,\Boldv$ represents the Mahalanobis norm.
The solution to eqn. \eqref{eqn:map_all} using all measurements is the \ac{EKF}  \ac{GNSS}-\ac{INS} integration. 
Given the solution $\delta \hat{\Boldx}^+_k $, the measurement corrected state is
$\Boldx^+ = \Boldx^- + \delta \Boldx^+$ (see the comment after eqn. \eqref{eqn:correction_est}).

To this point, we have assume that there are no outliers existing and all measurements have been used. 
The selection of measurements to avoid outliers is discussed next.

\subsection{Risk-Averse, Performance-Specified Estimation}
Traditional  measurement selection methods focus on outlier detection and exclusion. 
Alternatively, \ac{RAPS} endeavors to identify a subset of measurements that achieves a specified performance metric while minimizing the risk of outlier inclusion \cite{Aghapour_TCST_2019,hu2024optimization}.  

Measurement selection is achieved by introducing a non-binary vector $\Boldb \in \Rbb^m$ where $b_i \in [0,1]$ amplifies the noise of the $i$-th measurement:
$[\Boldgamma_k]_i\sim \mathcal{N}\left(\0, \frac{\sigma_i^2}{b_i^2}\right).$
 Integrating the selection vector $\Boldb$ into the \ac{MAP} cost function and omitting the subscript $k$, yields the risk  quantified  as
\begin{align}\
	C(\delta \Boldx, \Boldb) &= \norm{\delta\Boldx}^2_{\BoldP^-} + \norm{\BoldPsi(\Boldb)(\delta \Boldz - \BoldH\delta \Boldx)}^2_{\BoldR} \nonumber \\
	&= \norm{\delta\Boldx}^2_{\BoldP^-} + \sum_{i=1}^{m} \frac{\tau_i}{\sigma^2_i}(\delta z_i - \Boldh_i \delta \Boldx)^2
	\label{eqn:risk}
\end{align}
where $\BoldPsi(\Boldb) = diag(\Boldb)$, 
$\tau_i = b^2_i \in [0,1]$, 
$\delta z_i = [\delta \Boldz]_i$, and
$\Boldh_i$ is the $i$-th row of $\BoldH$. 

For a fixed value of $\Boldb$, the optimal posterior error state $\delta \Boldx$ solves the least squares problem:
\begin{align}\label{eqn:map_withb}
	\delta \hat{\Boldx}_{\Boldb}^+ = \,\underset{\delta \Boldx}{\text{argmin}} ~ C(\delta \Boldx, \Boldb)
\end{align}
with {\em posterior information matrix}  
\begin{align}\label{eqn:post_info}
	\BoldJ_{\Boldb}^+ = \BoldH^\top\,\BoldPsi(\Boldb)\, \BoldR^{-1}\,\BoldPsi(\Boldb)\, \BoldH + \BoldJ^-
\end{align}
where $\BoldJ^- = (\BoldP^-)^{-1}$ is the {\em prior information matrix} and $\BoldP_{\Boldb}^+ = \left( \BoldJ_{\Boldb}^+ \right)^{-1}$ is the {\em posterior error covariance matrix}.
Note that each choice of $\Boldb$ yields a potentially distinct posterior state estimate (i.e.,
$\Boldx^+ = \Boldx^- + \delta \Boldx_{\Boldb}^+$)
with a distinct predicted accuracy quantified by 
$\BoldP_{\Boldb}^+$ or $\left( \BoldJ_{\Boldb}^+ \right)^{-1}$.

Choosing ${\Boldb}$ to achieve a diagonal performance constraint with minimum risk yields the \ac{RAPS} optimization problem \cite{hu2024optimization, hu2024rapsppp} to be formalized as 
\begin{equation} \label{eqn:RAPS_nonbinary}
	\left.
	\begin{aligned}
		\delta \hat{\Boldx}^+ &\circeq \underset{\delta \Boldx^+, \Boldb}{\text{argmin}} ~ C(\delta \Boldx, \Boldb)\\ 
		&\text{s.t.:} \ \mbox{diag}(\BoldJ_{b}^+) \ge \BoldJ_l  \\
		&\ \ \ \ \ b_i \in [0,\,1] \mbox{ for } i\,=\,1,\,...\,m
	\end{aligned}
	\right\}
\end{equation}
where $\BoldJ_l$ represents a user-defined {\em information} vector with non-negative elements. 
The vector $\BoldJ_l$ corresponding to the lane-level specification (positioning accuracy better than 1 meter horizontally and 3 meters vertically, both at 95th percentile). Using Tchebycheff inequality \cite{Aghapour_TCST_2019}, \black
$\BoldJ_l \circeq [\BoldJ_p^\top, \BoldJ_v^\top]^\top$
with
$\BoldJ_p \circeq [1/0.05,\,1/0.05,\,1/0.45]^\top$
and 
$\BoldJ_v \circeq [1/0.025,\,0.025,\,1/0.225]^\top.$
The vectors $\BoldJ_p$ and $\BoldJ_v$ specify the  local tangent plane north, east, and vertical direction position and velocity information bounds.
Appendix A of \cite{hu2024optimization} shows that 
the information constraint $\mbox{diag}(\BoldJ_{b}^+) \ge \BoldJ_l $ can be expressed as the \ac{LMI}: $\BoldG \, \Boldb \ge \Boldd$ where
\begin{equation} \label{eqn:linear_constraint}
	\BoldG \circeq\begin{bmatrix}
		\frac{h_{11}^2}{\sigma_1^2} & \ldots & \frac{h_{m1}^2\T}{\sigma_m^2\B}  \\
		\vdots & \ddots & \vdots  \\
		\frac{h_{1n}^2}{\sigma_1^2} & \ldots & \frac{h_{mn}^2\T}{\sigma_m^2\B}  \\
	\end{bmatrix} \mbox{ and }	\Boldd \circeq \BoldJ_l - diag(\BoldJ^-).
\end{equation}

\subsection{Addressing Infeasible Epochs}
Both \ac{SAE} and lane-level \black specifications delineate performance requirements for open-sky conditions \cite{SAEJ2945,9043735}. 
However, in complex urban environments, such as deep urban canyons, the availability and reliability of satellite signals is markedly diminished. 
In such scenarios, the performance specification constraints may be unattainable at specific time instances, even if all available measurements are utilized, rendering the optimization problem as defined in \eqref{eqn:RAPS_nonbinary} infeasible. 

To resolve infeasible situations, soft-constrained \ac{RAPS} optimization  integrates slack variables. 
The formulation of the soft-constrained \ac{RAPS} approach is
\begin{align} \label{prob:RAPS_bcd_bi_slack}
	\left.
	\begin{aligned}
		\delta \hat{\Boldx}^+ & \circeq \,\underset{\delta \Boldx, \Boldb, \Boldmu}{\text{argmin}} ~C(\delta \Boldx, \Boldb) + \gamma \sum_{j=1}^{n}\mu_j \\ 
		\text{s.t.:} \ 
		&\Boldg_j \, \Boldb + \mu_j \geq d_j - \mathit{L}_j,  \mbox{ for } j\,=\,1,\,...\,n,\\
		&\mu_j \in \left\{
		\begin{aligned}
			&[0,\,\mathit{L}_j], ~ &\Boldg_j \, \1> d_j \\
			&[0,\,\Boldg_j \, \1],    &\Boldg_j \, \1 \leq d_j
		\end{aligned}
		\right.
		\\
		&b_i \in \{0,\,1\} \mbox{ for } i\,=\,1,\,...\,m.
	\end{aligned}
	\right\}
\end{align}
where 
$\mathit{L}_j = max(d_j - \Boldg_j\,\1,\,0)$,
$\Boldg_j$ is the $j$-th row of $\BoldG$, 
$d_j$ is the $j$-th element of $\Boldd$, and 
$\BoldJ_l(j)$ and $\BoldJ_d^-(j)$ represent the $j$-th element of $\BoldJ_l$ and $\BoldJ_d$, respectively.  
The soft constraint penalty term is $\gamma \, \sum_{j=1}^{n}\mu_j$, 
where the scaling parameter $\gamma$ is designed to modulate the influence of the slack variables.

When the performance constraints can be feasible (i.e., $\BoldG \, \1 \ge \Boldd$), Problem \eqref{prob:RAPS_bcd_bi_slack} simplifies to the original formulation in Problem \eqref{eqn:RAPS_nonbinary}. 
Conversely, when there is no feasible choice of $\Boldb$ (i.e., $\BoldG \, \1 < \Boldd$), the introduction of $\mu_j$ accommodates a reduction in the information to the maximum achievable level, thereby balancing  risk relative to  feasibility. 
The motivation for Problem \eqref{prob:RAPS_bcd_bi_slack} and its real-time feasible solution using block coordinate method is detailed in Sec. 6.5 of \cite{hu2024optimization}. 

\section{Outlier Accommodation for \ac{RTK} GNSS/\ac{INS} }\label{sec:estimation}
This section proposes a measurement update  framework  using \ac{GNSS} measurements for \ac{RTK}-\ac{INS} integration.

\subsection{\ac{GNSS} \ac{DD} Measurements}
\ac{GNSS} code and carrier phase measurements provide the means to estimate  absolute position, while Doppler measurements provide the means to estimate absolute velocity. 
Within a local vicinity, the \ac{RTK} approach performs a single-difference  between the rover and  base measurements to remove common-mode errors (e.g., ionosphere, troposphere, ephemeris, and satellite clock) and a second difference  between a pivot satellite and other satellites to eliminate receiver clock  and hardware bias errors.

The single-difference operation for satellite $s$ is 
\begin{align}
	\delta \rho^s &\circeq  \tilde\rho_r^s - 
		\left(\tilde\rho_b^s - R(\Boldp_b,\, \hat{\Boldp}_s) \right) \\
	\lambda \, \delta \phi^{s} &\circeq  \tilde\phi_r^s - 
		\left(\tilde\phi_b^s - R(\Boldp_b,\, \hat{\Boldp}_s) \right). 
\end{align}
where $\lambda$ is the wavelength and
$R(\Boldp_r, \Boldp_s) = \norm{\Boldp_r-\hat{\Boldp}_s}$ is the geometric range.
The symbols $\Boldp_r, \, \Boldp_s, \, \Boldp_b \in \Rbb^3$ denote the 
receiver, satellite, and base positions. 

These single-difference measurements are modeled as 
\begin{align}
	\delta \rho^{s} &= R(\Boldp_r,\, \hat{\Boldp}_s\,) + b_{r,\rho} + M_{\rho}^s +  \eta^s_{\rho}, 
	\label{eqn:sdcodemodel}\\
	\lambda \, \delta \phi^{s} &= R(\Boldp_r,\, \hat{\Boldp}_s) + b_{r,\phi} + \lambda \, N^s + M_{\phi}^s + \eta^s_{\phi}
	\label{eqn:sdphasemodel}
\end{align}
where 
 $b_{r,\rho}$ and $b_{r,\phi}$ represent the receiver clock bias for pseudorange and phase; 
$N^s$ is the integer ambiguity;
$M_{\rho}^s$ and $M_{\phi}^s$ represent multipath and \ac{NLOS} errors; and
$\eta^s_{\rho}$ and $\eta^s_{\phi}$ represent measurement noise. 
The combined multipath and measurement noise is assumed to be white with a Gaussian distribution, represented as $(M^s+\eta^s) \sim \mathcal{N}(0, \sigma^s)$, where $\sigma^s$ depends on the satellite elevation.
Carrier phase noise \ac{STD} is at the millimeter level, with its multipath error usually less than a few centimeters \cite{georgiadou1988carrier}.
For satellites at  high elevations and in open-sky conditions, code noise \ac{STD} is at the decimeter level, with its associated multipath error typically under a few meters.

Denoting the pivot satellite by $o$, the \ac{DD} code and phase measurements models are  (see Section 8.83 in \cite{farrell2008aided})
\begin{align}
	\bigtriangledown \rho^{s} &= R(\Boldp_r, \Boldp_s) - R(\Boldp_r, \Boldp_o) + M_{\rho}^{s,o} + \eta^{s,o}_{\rho},\\
	\lambda \, \bigtriangledown \phi^{s} &= R(\Boldp_r, \Boldp_s) - R(\Boldp_r, \Boldp_o) + \lambda \, N^{s,o} + M_{\phi}^{s,o} + \eta^{s,o}_{\phi}
\end{align}
where $M^{s,o} = M^s-M^o$, $N^{s,o} = N^s-N^o$ is the DD ambiguity, and $\eta^{s,o} = \eta^s - \eta^o$.
The linearized code and phase residual models are 
\begin{align}
	\delta z_{s, \rho} &= (\1_r^s - \1_r^o) \, \Boldp_r + M_{\rho}^{s,o} + \eta^{s,o}_{\rho}, \\
	\lambda \, \delta z_{s, \phi} &= (\1_s^r - \1_o^r) \, \Boldp_r + \lambda N^{s,o} + M_{\phi}^{s,o} + \eta^{s,o}_{\rho}.
\end{align}

After compensating for satellite velocity and clock drift, the single-differenced Doppler measurement model is
\begin{align}\label{eqn:sd_dop}
	\delta D^s &= \1_r^s \, \Boldv_r +\dot{b}_r + \eta^s_D
\end{align}
where $\1_r^s = \frac{\Boldp_r^- -\hat{\Boldp}_s}{\|\Boldp_r^- -\hat{\Boldp}_s \|}$ is the line-of-sight vector from the satellite to the receiver,
$\Boldp_r^-$ is the prior position of the receiver, 
$\Boldv_r$ is the receiver velocity vector, and 
$\eta^s_D \sim \mathcal{N}(0, \sigma_{s,D})$ is the Doppler measurement noise. 
The \ac{DD} Doppler measurement model for satellite $s$ is 
\begin{align}\label{eqn:linear_dop}
	\bigtriangledown D^s &= (\1_r^s - \1_r^o)^\top \, \Boldv_r + \eta^{s,o}_D
\end{align}
where $\eta^{s,o}_D = \eta^s_D-\eta^o_D$.

\subsection{Outliers}
Outliers in GNSS are due to multipath, non-line-of-sight, and other anomalies for which there are no explicit models. 
Outliers are defined as measurements that are unlikely based on the measurement model and prior information; therefore, they have high risk as quantified by eqn. \eqref{eqn:risk}. 
The Gaussian noise model assumption stated after eqn. \eqref{eqn:sdphasemodel} is  reasonable under open-sky scenarios, where code multipath errors are generally minor and tend to increase as elevation decreases \cite{khanafseh2018gnss,uwineza2019characterizing}. 
However, in dense urban environments, code multipath and \ac{NLOS} effects can escalate to tens of meters  \cite{hsu2017gnss}. 
Thus, when these errors become significant relative to $\sigma^s$, they are treated as outliers.

\subsection{Measurement Residuals}
At epoch $t_k$, let $\Boldz^s = [\bigtriangledown \rho^{s}, \bigtriangledown D^s, \lambda \, \bigtriangledown \phi^{s}]^\top$ denote the vector of measurements for  satellite $s$, which can be modeled as
\begin{equation}
	\Boldz^s = \Boldh^s(\Boldchi) + \Boldepsilon^s,
\end{equation}
where  $\Boldchi = [\Boldx; \Boldx_{N}]$ and
$\Boldx_{N} \in \Rbb^m$ denotes a vector of \ac{DD} ambiguities (i.e., $N_{s,o}$)
\begin{equation}
	\Boldh^s(\Boldchi) = \begin{bmatrix} R(\Boldp_r, \Boldp_s) - R(\Boldp_r, \Boldp_o) \\ R(\Boldp_r, \Boldp_s) - R(\Boldp_r, \Boldp_o) + \lambda \, N^{s,o} \\ (\1_r^s - \1_r^o)^\top \, \Boldv_r \end{bmatrix},
\end{equation}
and the noise vector $\Boldepsilon^s \sim \mathcal{N}(0, \BoldR_s)$ with 
\begin{align}
	\BoldR_s = \begin{bmatrix} \sigma^2_{s,\rho} + \sigma^2_{o,\rho} & 0 & 0 \\ 
		0 & \sigma^2_{s,D} + \sigma^2_{o,D} & 0 \\
		0 & 0 & \sigma^2_{s,\phi} + \sigma^2_{o,\phi}
	\end{bmatrix}.
\end{align}
In this paper, the noise matrix is assumed to be uncorrelated between satellites. 
This assumption is not strictly true in practice due to the double-difference operation.

The predicted measurement and measurement residual are computed as
\begin{equation}
	\hat{\Boldz}^s \circeq \Boldh^s(\hat{\Boldchi}^-) \text{ and } 
	\delta \Boldz^s \circeq \Boldz^s - \hat{\Boldz}^s,
\end{equation}
with linearized residual modeled as
\begin{equation}
	\delta \mathbf{z}^s = \BoldH_s \, \delta \Boldchi + \gamma^s
\end{equation}
where $\delta \Boldchi = \Boldchi - \hat{\Boldchi}^-$, and the measurement matrix is
\begin{align}
	\BoldH_s = 
	\begin{bmatrix}
		\Boldh_{\rho}^s \\ \Boldh_{D}^s \\ \Boldh_{\phi}^s
	\end{bmatrix}
	= \begin{bmatrix}
		(\1_r^s - \1_r^o) & \0 & \0 & \0 \\ 
		\0 & (\1_r^s - \1_r^o) & \0 & \0 \\
		(\1_r^s - \1_r^o) & \0 & \0 & \Bolde_s
	\end{bmatrix}
\end{align}
where $\0$  is a conformal matrix containing all zeros, and
$\Bolde_s$ represents $s$-th row of the $m\times m$ identity matrix.

The concatenation of the residuals per satellite yields
\begin{equation}
	\Boldz = \BoldH \, \delta \Boldchi + \Boldepsilon
\end{equation}
where $\BoldH$ and $\Boldepsilon$ are the concatenation of $\BoldH_s$ and $\Boldepsilon_s$.

The models presented herein assume that  the antenna and IMU are co-located. 
When there is significant separation between the \ac{IMU} and the GNSS antenna,  lever arm compensation is required (see Sec. 11.8  in \cite{farrell2008aided}).

\subsection{Risk-Averse Optimization-based \ac{RTK} Float Solution} \label{sec:rtk-raps}
A critical aspect of using carrier phase measurements is the estimation of integer phase ambiguities. 
In open-sky conditions the integer ambiguities are constant over many epochs.
In deep-urban canyons, deleterious  effects (e.g., multipath) can lead to temporary degradation in the quality of \ac{GNSS} code measurements and changes to the integer ambiguities. 
Traditional cycle slip detection and integer estimation methods may fail to detect or re-determine the integer ambiguities in real-time. 
Therefore, herein,  we employ the {\em instantaneous \ac{RTK} method} \cite{parkins2011increasing,odolinski2015combined} in which integer ambiguities are re-initialized at every epoch without prior information (i.e., infinite prior covariance).
This approach is insensitive to cycle slips and signal anomalies, it relies solely on the observations from a single epoch.
The trade-off is that, during intervals when the integers are constant, the system does not get the advantage of their prior knowledge.

\begin{figure}[b]
	\centering
	\includegraphics[width=\linewidth]{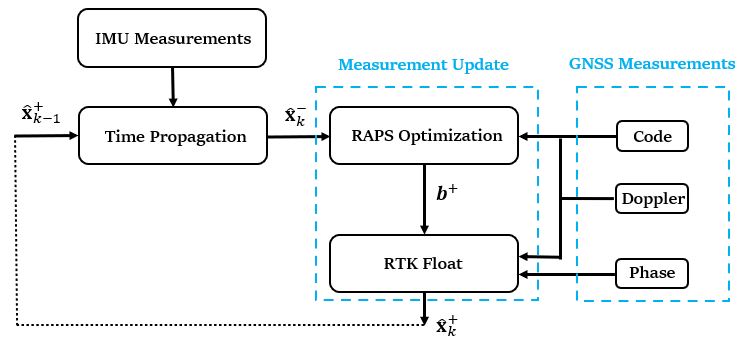}
	\caption{The  \ac{RAPS} \ac{RTK} \ac{GNSS}/\ac{INS} workflow. }
	\label{fig:workflow}
\end{figure}

In \ac{RTK} instantaneous, the prior integer ambiguities are unknown and the parameter $\sigma_{s,\phi}$ is small; therefore,  
the cost term $(\lambda \, \delta z_{s, \phi} - \Boldh^s_{\phi} \delta \Boldchi)/\sigma_{s,\phi}$ and 
the information term $((\Boldh^s_{\phi})^\top \Boldh^s_{\phi})/\sigma^2_{s,\phi}$  associated with phase measurement are large. 
Also, the magnitude of the phase multipath error is shown in \cite{georgiadou1988carrier} to be less than $\lambda/4$.
Larger multipath may result in a cycle slip, but the  \ac{RTK} instantaneous method is immune to cycle slips.
Therefore, the phase measurement is effectively immune from outliers.
Because the ambiguity estimation relies on both the code and phase data, the method cannot rely solely on phase measurements; therefore,
measurement selection is based on the code and Doppler measurements, which are affected by outliers.
Then, once the vector $\Boldb$ is determined, it is applied to both the code and phase range measurements for each satellite.
After $\Boldb$ is determined, the state is estimated from the \ac{RTK} float solutions. 
Fig. \ref{fig:workflow} illustrates the workflow.

After obtaining the prior vehicle state information from the time propagation, the \ac{RAPS} \ac{RTK} \ac{GNSS}/\ac{INS} process is executed in two steps:
\begin{description}
	\item[1.] \ac{RAPS} Measurement Selection:
	The \ac{RAPS} optimization problem is as formulated in Problem \eqref{prob:RAPS_bcd_bi_slack} with cost function  defined as:
	\begin{align}
		C(\Boldx, \bar\Boldb, \mu) = &\norm{\delta\Boldx}^2_{\BoldP^-} + \nonumber \\
		&\sum_{s=1}^{m} \frac{\bar\Boldb_s}{\sigma^2_s}(\delta \bar\Boldz_s - \bar\BoldH_s\delta \Boldchi)^2 + \gamma \sum_{j=1}^{n}\Boldmu_j
	\end{align}
	where $\bar\Boldb_s = [\Boldb_\rho, \Boldb_D]$,  $\delta \bar\Boldz_s$, and $\bar\BoldH_s$ denote the  measurement selection vector, residuals, and the corresponding rows of the measurement matrix for the code  and Doppler measurements, respectively. 
	The constraints from the original problem formulation remain unchanged. 
	The optimization aims to derive the optimal solution $\bar\Boldb^{\ast}$.
	While the optimal $\Boldchi^\ast$ and $\Boldmu^\ast$ based on code and Doppler are also estimated, they are discard.
	
	\item[2.] Find the \ac{RTK} float solution using the given $\bar\Boldb$: 
	In this phase, we define $\Boldb = [\Boldb_\rho, \Boldb_D, \Boldb_\phi]$ where $\Boldb_\phi = \Boldb_\rho$. 
	This value of $\Boldb$ is fixed.
	The posterior error state estimate is calculated as:
	\begin{align}\label{eqn:rtk_map}
		\delta \hat{\Boldchi}^+ & = \underset{\delta \Boldchi}{\text{argmin}} \, \Big(\norm{\delta\Boldx}^2_{\BoldP^-} + 
		\norm{\BoldPsi(\Boldb)(\delta \Boldz - \BoldH \, \delta \Boldchi)}^2_{\BoldR^{-1}} \Big) 
		\nonumber
	\end{align}
	which includes the phase measurements with their ambiguities estimated as real numbers. 
	The posterior state is obtained from $\Boldx^+ = \Boldx^- + \delta \Boldx^+$. The posterior information can be computed from eqn. \eqref{eqn:post_info}. 
\end{description}

\section{Experiment Performance and Analysis}\label{sec:exp}

The experimental analysis employs the open-source TEX-CUP dataset \cite{narula2020tex}. 
The \ac{RTK}-\ac{INS} integration utilizes single-frequency measurements from GPS L1, GLONASS L1, GALILEO E1, and Beidou B1 at 1 Hz measurement epochs. 
The elevation cut-off is set at 10 degrees. 
The inertial measurements are provided by a low-cost smartphone-grade Bosch BMX055 \ac{IMU} with a sampling rate of 150 Hz.
\ac{GNSS} data were acquired using dual Septentrio receivers—one for the rover and one for the base station. 
Separately, the data set includes a ground truth trajectory for evaluation purposes.

The experimental route traversing areas within the west campus of The University of Texas at Austin and downtown Austin,  containing  viaducts, high-rise buildings, and dense foliage; therefore, it presents a challenging urban environment. 
Such areas are particularly susceptible to frequent and substantial GNSS outliers due to severe multipath effects and \ac{NLOS} errors for any satellites. 
The data set spans approximately 1.5 hours, including about 480 seconds of stationary status in an open-sky environment. 
Visual street views of deep-urban areas are illustrated in Fig. 5 of \cite{narula2020tex}.

Previous research has established the effectiveness, robustness, and real-time capabilities of \ac{RAPS} using a non-binary measurement selection vector $\Boldb$ in Differential \ac{GNSS} applications. 
Compared to traditional methods, \ac{RAPS} selects a subset of measurements to achieve a specified performance with minimum risk. 
The avaliablity of GNSS measurements for this dataset has been illustrated in Fig. 7.3 of \cite{hu2024optimization}.
This experiment focuses on the positioning performance in \ac{RTK}-Aided \ac{INS} applications.
The estimation methods that will be compared are:
\begin{description}
	\item[RAPS-PVA:] 
	This approach propagates the state using a standard \ac{PVA} model. For the  \ac{RAPS} 
	measurement update,  only code and Doppler measurements are used.
	Results presented in \cite{hu2024optimization} for the same dataset are used for comparison here.  
	
	\item[RAPS-PVA-RTK:] Time propagation is again performed using the \ac{PVA} approach. 
	The posterior state estimate is solved by the proposed \ac{RAPS} \ac{RTK} \ac{GNSS} strategy described in Sec. \ref{sec:rtk-raps}.

	\item[EKF-INS-RTK:] The time propagation is performed using the  \ac{IMU}  measurements and the INS methods described in Section \ref{sect:INS}. 
	The posterior state is estimated by minimizing the cost function as defined in eqn. \eqref{eqn:map_all} using all the measurements including carrier phase. 
	
	\item[TD-INS-RTK:] The time propagation is performed using  the  \ac{IMU}  measurements and the INS methods described in Section \ref{sect:INS}. 
	This approach selects $\Boldb$ based on the \ac{TD} method with a decision parameter equal to 2 (see \cite{hu2024rapsppp}).
	It then determine the posterior state using eqn. \eqref{eqn:map_withb}.

	\item[RAPS-INS-RTK:] The time propagation is performed using  the  \ac{IMU}  measurements and the INS methods described in Section \ref{sect:INS}. 
	The posterior state estimate is solved via the \ac{RAPS}-\ac{RTK} strategy proposed in Sec. \ref{sec:rtk-raps}.
\end{description}
For all the RAPS methods, $\gamma$ has been set to $50$, providing a balanced trade-off between constraint satisfaction and optimization robustness (see Sec. 7.2 of  \cite{hu2024optimization}).

\begin{table}[tb]
	\centering
	\caption{Horizontal position error statistics.}
	\begin{tabular}{c|ccccc}
		\hline
		\multirow{2}{*}{Methods}\T & \multirow{2}{*}{\begin{tabular}[c]{@{}c@{}}Mean\\ (m)\end{tabular}} & \multirow{2}{*}{\begin{tabular}[c]{@{}c@{}}RMS\\ (m)\end{tabular}} & \multirow{2}{*}{\begin{tabular}[c]{@{}c@{}}Max\\ (m)\end{tabular}} & \multirow{2}{*}{$\leq$ 1.0 m} & \multirow{2}{*}{$\leq$ 1.5 m} \\
		&     &     &    &    &   \T \\ \hline
		RAPS-PVA \cite{hu2024optimization}\T      &  4.23  & 9.73 & 56.22 & 60.22\% & 67.97\% \\
		RAPS-PVA-RTK\T      &  3.43  & 10.84 & 57.95 & 73.24\% & 79.53\% \\
		EKF-INS-RTK\T     &  3.36 & 11.42 & 163.67 & 67.76\% & 73.24\% \\
		TD-INS-RTK\T    &  2.90  & 9.97   &  177.23 & 68.03\% & 73.87\% \\
		RAPS-INS-RTK \T       &   \textbf{0.92}  & \textbf{1.81} & \textbf{16.44} & \textbf{77.80\%} & \textbf{85.84\%} \\ \hline
	\end{tabular}
	\label{tab:status_hor}
\end{table}

\begin{table}[tb]
	\centering
	\caption{Vertical position error statistics.}
	\begin{tabular}{c|cccc}
		\hline
		\multirow{2}{*}{Methods}\T & \multirow{2}{*}{\begin{tabular}[c]{@{}c@{}}Mean\\ (m)\end{tabular}} & \multirow{2}{*}{\begin{tabular}[c]{@{}c@{}}RMS\\ (m)\end{tabular}} & \multirow{2}{*}{\begin{tabular}[c]{@{}c@{}}Max\\ (m)\end{tabular}} & \multirow{2}{*}{$\leq$ 3.0 m}\\
		&     &     &    \T \\ \hline
		RAPS-PVA \cite{hu2024optimization}\T      &  7.51  & 18.88 & 131.93 & 68.59\% \\
		RAPS-PVA-RTK\T      &  5.94  & 19.01 & 109.46 & 85.92\% \\
		EKF-INS-RTK\T     & 5.61 & 17.72 & 310.16 & 76.85\% \\
		TD-INS-RTK\T    &  4.74  & 14.64  &  243.51 & 77.78\% \\
		RAPS-INS-RTK \T       &   \textbf{1.69}  & \textbf{4.67} & \textbf{53.12} & \textbf{92.07\%} \\ \hline
	\end{tabular}
	\label{tab:status_ver}
\end{table}

Tables \ref{tab:status_hor} and \ref{tab:status_ver} summarize the horizontal and vertical positioning performance statistics for each estimation method. 
The best results are highlighted in bold. The \ac{SAE} specification requires positioning accuracy better than 1.5 meters horizontally and 3 meters vertically, both at 68\% confidence. 
The tables include results for horizontal errors less than 1 meter and 1.5 meters, and vertical errors less than 3 meters.
	
Comparing RAPS-PVA-RTK with the RAPS-PVA results, RAPS-PVA-RTK exhibits 13\% and 17\% improvements in horizontal and vertical performance, respectively. This demonstrates the benefits from using the significantly more accurate carrier phase measurements.

In the experiments using RTK-Aided \ac{INS}, \ac{TD} delivers slight improvements relative to the \ac{EKF}. 
These results align with prior research where \ac{TD} has the capability to remove severe outliers, but does not achieve the minimum risk and  does not consider the performance specification.  
During the state estimation, the ambiguity estimation relies on both the code and phase data. 
When significant outliers from code measurements are not rejected, those effects ultimately lead to corrupted position and ambiguity state estimates.
In contrast, \ac{RAPS} optimally minimizes the risk by selecting a subset of measurements when the performance specifications are feasible or alternatively adopts a soft constraint when strict specification is not achievable \cite{hu2024optimization}. 
When a code measurement is not selected, RAPS also discards the phase measurement for the same satellite.
The experiment result showcase that \ac{RAPS} performs about 10\% better than \ac{EKF} and \ac{TD}. 
The maximum error is also significantly reduced to 16.44 meters and has the lowest mean and \ac{RMS} errors.

\begin{figure}[bt]
	\centering{
	\includegraphics[width=\linewidth]{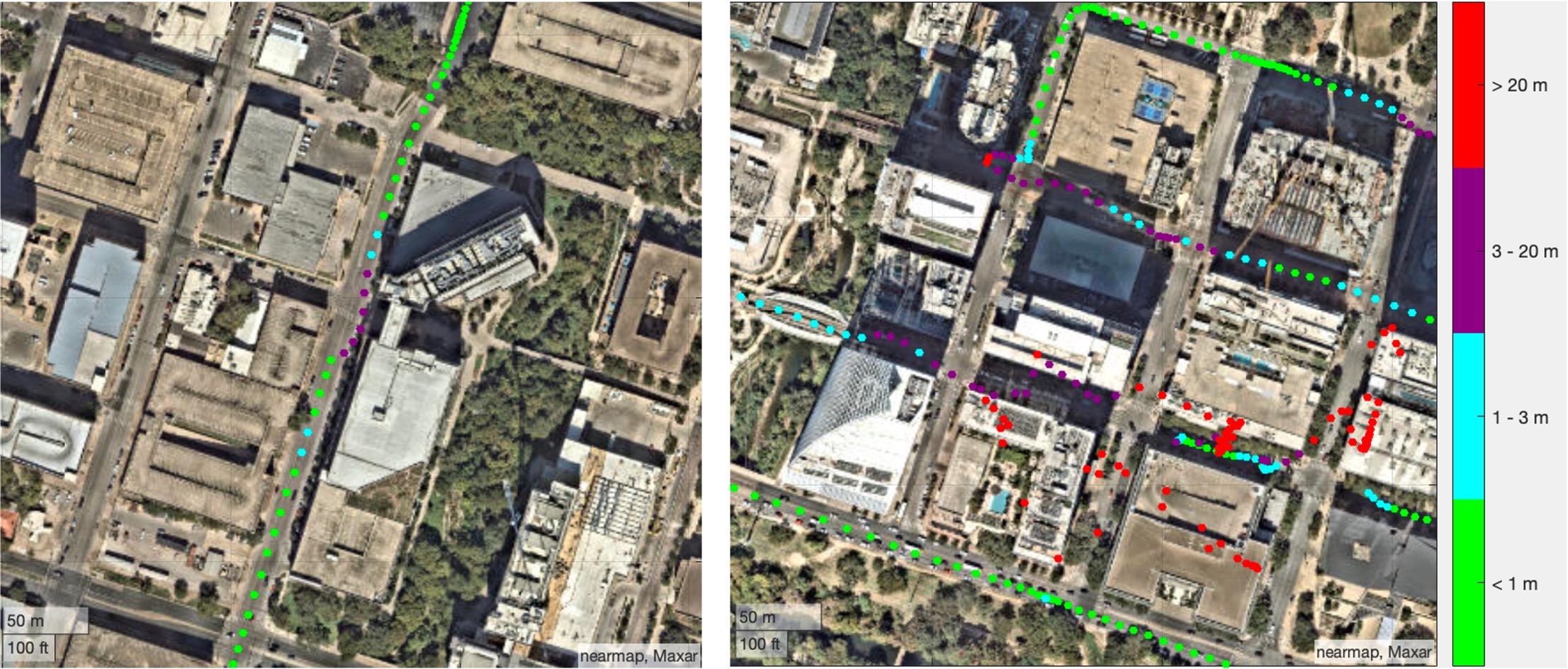}
	\caption{Trajectory estimated by TD-INS-RTK with horizontal error magnitude identified by color.}
	\label{fig:sat_view_td}}
\vspace{0.1in}
	\centering{
	\includegraphics[width=\linewidth]{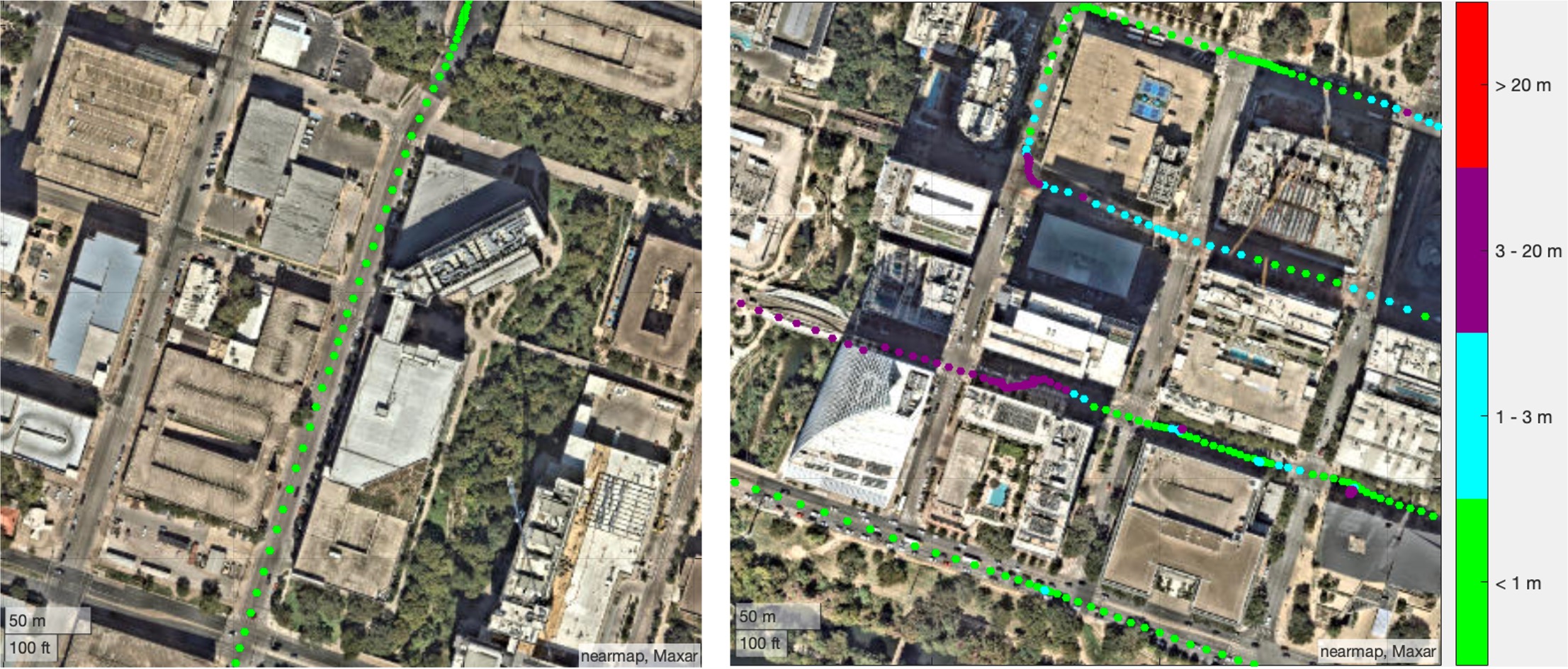}
	\caption{Trajectory estimated by RAPS-INS-RTK with horizontal error magnitude identified by color.}
	\label{fig:sat_view_raps}}
\end{figure}

Fig. \ref{fig:sat_view_td} and \ref{fig:sat_view_raps} illustrate the trajectories estimated by TD-INS-RTK and RAPS-INS-RTK, respectively. 
Each point represents the estimated position at a \ac{GNSS} measurement epoch, marked by a color  to indicate its horizontal accuracy. 
The error ranges are depicted in the color bar on the right side.
The left panels of the figures present the route passing the Dell Medical School buildings.
The left panels demonstrate that RAPS-INS-RTK maintains decimeter-level performance, whereas TD-INS-RTK yields positioning errors of 3-20 m. 
The right panels display the path through a deep-urban area near The Sailboat Building, where multiple skyscrapers surround the site, causing severe \ac{GNSS}  errors.
TD-INS-RTK exhibits large errors due to missed outlier detections causing inconsistencies between the actual and estimated mean and covariance. 
Note that the red points indicate errors greater than 20 meters. These errors are clearly visible because the points are located in the top of buildings instead of the road.
The estimator is overconfident in the corrupted state estimate, thus rendering subsequent outlier decisions unreliable.
In contrast, \ac{RAPS} selects a subset of measurements to minimize outlier risk. It delivers smoother real-time estimation and maintains road-level accuracy as the vehicle navigates areas surrounded by skyscrapers.

The statistical results  underscore the benefits of:
\begin{description}
	\item[IMU:]  RAPS INS  outperforms the RAPS PVA  due to the additional available sensor information. 
	\item[RAPS:] Optimal measurement selection results in better consistency between the estimated mean and its error covariance matrix. 
\end{description}
The trade-off in using  INS, instead of PVA, is that an additional sensor and computation are required. 
The trade-off in using  RAPS, instead of either TD or EKF, is that additional  computation is required.

\section{Conclusion and Discussion}\label{sec:conclu}
In urban settings, \ac{GNSS} is often subject to substantial errors from multipath effects and non-line-of-sight effects.
With the goal of tackling such urban navigation challenges, 
this article introduced a framework integrating \ac{RAPS} with tightly coupled \ac{RTK} aided \ac{INS}.
The approach is innovative in its strategy to address outliers in \ac{GNSS} measurements and in its approach to incorporate high-accuracy carrier phase measurements.
Our proposed RTK-INS-RAPS framework leverages a non-binary measurement selection vector that de-weights a subset of satellite code and phase measurements to minimizing outlier risk while ensuring that a performance specification is achieved when it is feasible. 
Experimental results using the open-source TEX-CUP data set underscore the effectiveness of this method, achieving 85.84\% of horizontal errors below 1.5 meters and 92.07\% of vertical errors below 3 meters. 
This performance not only exceeds \ac{SAE} standards, but also shows a 10\% enhancement compared to traditional methods like the \ac{EKF} and threshold decision.

There are several compelling avenues for further research. 
This paper centers on using carrier phase measurements within \ac{RAPS} to achieve \ac{RTK} in float integer ambiguities. 
If the float ambiguity estimates can be fixed to the correct integers, then
centimeter-level accuracy will be achieved.
Correctly resolving ambiguities as integers requires additional refinements and integrity tests. 
Moreover, adding additional sensor modalities, such as LIDAR and vision \cite{liu2024td3,jin2023visual,li-19-segmentation}, will improve the reliability and accuracy of the navigation system under diverse environments such as tunnels, areas surrounding skyscrapers, and under bridges. 

\section{Acknowledgment}
This article is based upon work supported by USDOT CARNATIONS (Grant No. 69A3552348324), NSF award number 2312395,  and  the UCR KA Endowment. Any opinions, findings, and conclusions or recommendations expressed in this publication are those of the authors and do not necessarily reflect the views of the sponsors. Any errors or omissions are the responsibility of the authors.

\bibliographystyle{Biblio/IEEEtran.bst}
\bibliography{Biblio/IEEEabrv.bib,Biblio/References.bib}

\end{document}